\documentclass[conference]{IEEEtran}
\IEEEoverridecommandlockouts

\usepackage{cite}
\usepackage{amsmath,amssymb,amsfonts}
\usepackage{algorithmic}
\usepackage{graphicx}
\usepackage{textcomp}
\usepackage{xcolor}
\usepackage{float}
\usepackage{stfloats}
\usepackage{multirow}
\usepackage{balance}
\usepackage{booktabs}  
\def\BibTeX{{\rm B\kern-.05em{\sc i\kern-.025em b}\kern-.08em
    T\kern-.1667em\lower.7ex\hbox{E}\kern-.125emX}}
\begin{document}

\title{Single-Channel Tissue Segmentation via Cross-Modal
Distillation from Foundation Models}

\author{\IEEEauthorblockN{Anonymous Authors}
}


\author{
\begin{tabular}{cc}
\begin{minipage}[t]{0.45\textwidth}\centering
\text{1\textsuperscript{st} Sakib Mohammad}\\
\textit{Department of Engineering Technology}\\
\textit{Fairmont State University}\\
Fairmont, WV, USA\\
smohammad@fairmontstate.edu
\vspace{1.2em}
\end{minipage}
&
\begin{minipage}[t]{0.45\textwidth}\centering
\text{2\textsuperscript{nd} Jarin Ritu}\\
\textit{Department of Electrical and Computer Engineering}\\
\textit{Texas A\&M University}\\
College Station, TX, USA\\
jarin.ritu@tamu.edu
\end{minipage}
\\[2.7em]
\multicolumn{2}{c}{
\begin{minipage}[t]{0.45\textwidth}\centering
\text{3\textsuperscript{rd} Md Sakhawat Hossain}\\
\textit{Department of Mechanical Engineering}\\
\textit{Auburn University}\\
Auburn, AL, USA\\
mdh0127@auburn.edu
\end{minipage}
}
\end{tabular}
}
\maketitle


\begin{abstract}
Multiplexed fluorescence microscopy improves tissue
segmentation by providing complementary channels including
nuclear (DAPI) and membrane (E-cadherin), that
together encode richer spatial context than
single-channel imaging alone. However, multiplexed
models require all channels at inference, limiting
deployment where only a subset is available. This
work proposes a cross-modal knowledge distillation
framework that transfers semantic information from
a frozen foundation model teacher processing
multiplexed input to a lightweight student operating
on the nuclear channel only. The distillation
objective combines MSE-based probability matching,
boundary-aware supervision, and learnable uncertainty
weighting. SAM ViT-H and CellSAM are evaluated as
teachers across four U-Net students: Swin-Tiny
(27M), ResNet18 (11M), EfficientNet-B0 (5.3M), and
MobileNetV3 (1.5M), on TissueNet and BBBC038. On
TissueNet, the SAM-distilled Swin-Tiny student
achieves Dice 78.36 ($\pm$1.44), a 13.05-point
improvement over the no-KD baseline (65.31
$\pm$1.35) and 87.9\% recovery of teacher oracle
performance (89.12 $\pm$1.21) at a $23\times$
parameter reduction. KD consistently improves all
four students by $\sim$12 Dice points, confirming
architecture-agnostic distillation. SAM ViT-H
outperforms CellSAM as teacher across all settings.
Cross-dataset evaluation on BBBC038 shows consistent
gains without teacher retraining.
\end{abstract}

\section{Introduction}

Accurate segmentation of tissues and cells is a fundamental
task in biomedical image analysis, supporting applications
such as disease characterization, spatial profiling, and
quantitative pathology~\cite{moen2019deep}. Despite
significant advances in deep learning based
segmentation~\cite{ronneberger2015unet}, accurate
delineation of cell boundaries remains challenging due to
variations in cell morphology, imaging modalities, and
tissue heterogeneity~\cite{greenwald2022wholecell}. Recent advances in multiplexed imaging have enabled the
acquisition of high dimensional data where multiple
image channels capture complementary biological
features within a single sample, for example, nuclear
staining (DAPI) and membrane markers (E-cadherin) that
together provide a richer spatial context than either
channel alone. Models trained on such
representations consistently outperform single-channel
counterparts on segmentation
benchmarks~\cite{greenwald2022wholecell}. 

However, multiplexed imaging increases memory use, latency, and model complexity, limiting scalability in resource-constrained settings. It also requires all channels at inference, which may be impractical in real-life scenarios where only a single channel may be available. These constraints motivate methods that can reduce input dimensionality while preserving the semantic information encoded in multiplexed representations. Knowledge distillation (KD)~\cite{hinton2015distilling} offers a principled solution: a large teacher model trained on high dimensional inputs transfer implicit knowledge to a compact student model. While KD has been widely applied to model compression in classification~\cite{hinton2015distilling}
and semantic segmentation~\cite{liu2019structured}, its
application to \textit{cross-modal} distillation,
where the teacher and student operate on inputs of
different dimensionality, remains largely unexplored,
particularly in biomedical imaging.

In this work, we propose a cross-modal KD framework that transfers semantic information from multiplexed microscopy images to single-channel inputs for tissue segmentation. A frozen foundation-model teacher processes nuclear and membrane channels and generates soft-targets to guide a lightweight student trained only on the nuclear channel.
We evaluate two teachers:
Segment Anything Model (SAM)~\cite{kirillov2023sam} with a ViT-H~\cite{Dosovitskiy2020AnII}, and
CellSAM~\cite{israel2023cellsam}, and four U-Net~\cite{ronneberger2015unet} encoder backbones: Swin-Tiny~\cite{liu2021swin}, ResNet18~\cite{he2016resnet}, EfficientNet-B0~\cite{tan2019efficientnet}, and MobileNetV3~\cite{Howard2019SearchingFM}. The framework is further assessed on BBBC038~\cite{caicedo2019nucleus} to examine cross-dataset generalization without teacher retraining
The main contributions of this work are as follows:

\begin{itemize}
    \item Cross-modal knowledge distillation that transfers semantic information from a frozen multiplexed foundation model teacher to a lightweight single-channel student for tissue segmentation under missing-channel settings

    \item An uncertainty-weighted distillation objective that integrates soft-target matching and boundary-aware supervision to balance heterogeneous training signals

    \item Extensive evaluation of SAM ViT-H and CellSAM teachers with four student architectures on TissueNet and BBBC038, achieving consistent performance improvements and up to $421\times$ parameter reduction

    \item A cross-dataset distillation protocol that leverages TissueNet teacher predictions to supervise students trained exclusively on BBBC038, without teacher retraining or target-domain multiplexed data
\end{itemize}
\begin{figure*}[!htbp]
    \centering
    \includegraphics[width=0.9\linewidth]{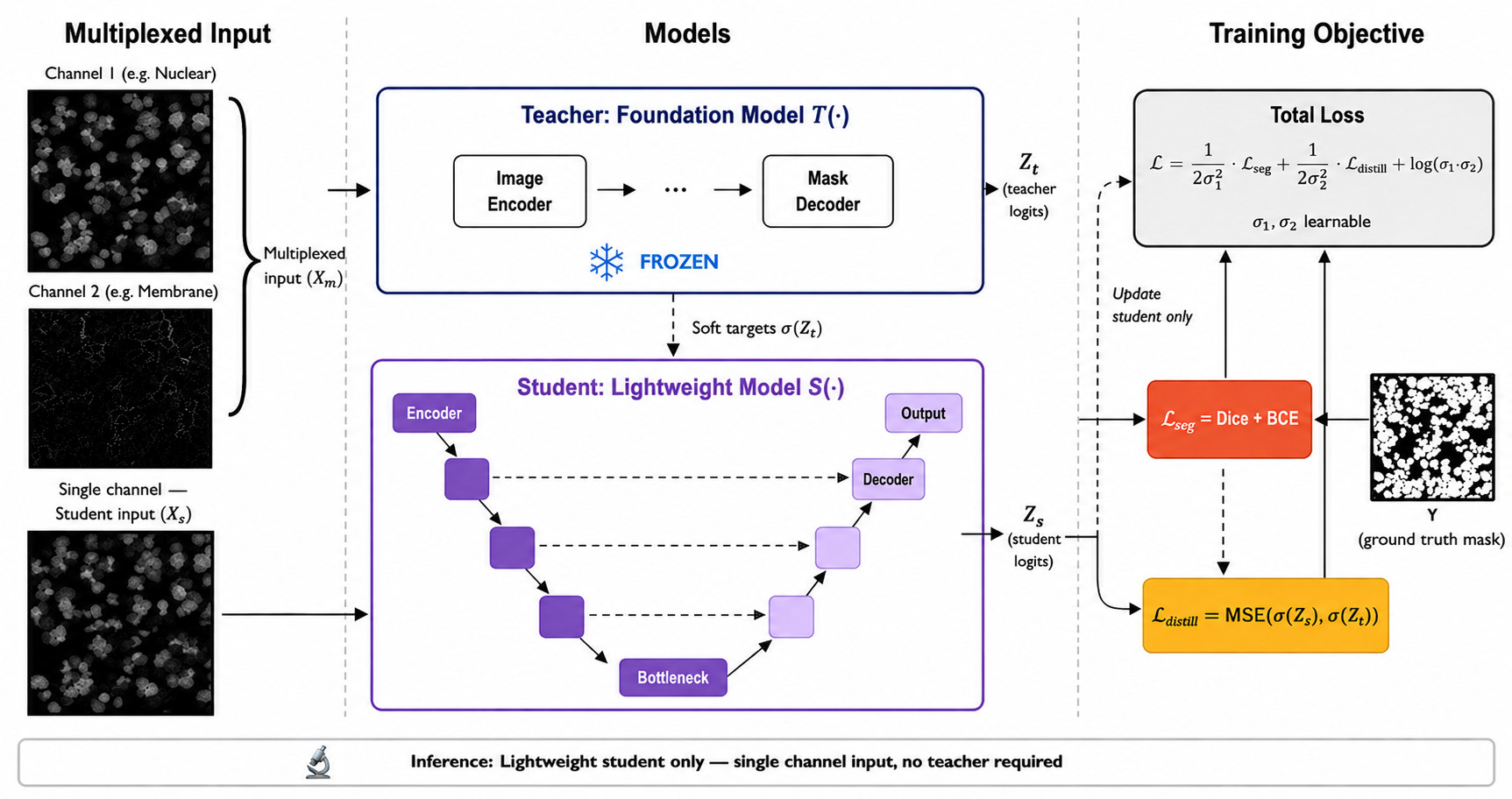}
    \caption{Overview of the proposed cross-modal knowledge
    distillation framework. A frozen foundation model
    teacher $T(\cdot)$ processes multiplexed images $X_m$
    and generates soft targets $\sigma(Z_t)$. A lightweight
    student model $S(\cdot)$ is trained on single-channel
    input $X_s$ using a combination of segmentation loss
    $\mathcal{L}_{seg}$, distillation loss
    $\mathcal{L}_{distill}$, and learnable uncertainty
    weights $\sigma_1,
    \sigma_2$~\cite{kendall2018}. At inference, only the
    student is deployed — no teacher or multiplexed input
    is required.}
    \label{fig:methodology}
\end{figure*}
\subsection{Related Work}

\subsubsection{Biomedical Image Segmentation}

Deep learning has substantially advanced cell and tissue segmentation, with U-Net~\cite{ronneberger2015unet} becoming a standard architecture due to its encoder--decoder design and skip connections. More recent approaches have incorporated transformer-based encoders such as Swin Transformer~\cite{liu2021swin} and lightweight backbones including MobileNet~\cite{Howard2017MobileNetsEC} and EfficientNet~\cite{tan2019efficientnet}. Large-scale benchmarks such as TissueNet~\cite{greenwald2022wholecell} have further shown that combining nuclear and membrane channels improves segmentation performance over single-channel imaging, motivating the transfer of multiplexed information to single-channel models.

\subsubsection{Foundation Models for Segmentation}

Foundation models have recently demonstrated strong zero-shot segmentation capabilities. SAM~\cite{kirillov2023sam} established a general-purpose segmentation framework, while MedSAM~\cite{ma2024medsam} and CellSAM~\cite{israel2023cellsam} adapted these capabilities to medical and cellular imaging domains. Although these models achieve strong segmentation performance, they typically require the same input modalities available during training and inference. In this work, SAM and CellSAM are used as frozen teachers to transfer multiplexed information to single-channel students.

\subsubsection{Knowledge Distillation}

Knowledge distillation (KD)~\cite{hinton2015distilling} transfers knowledge from a high-capacity teacher to a compact student using soft prediction targets and has been widely applied to model compression in classification and dense prediction tasks~\cite{liu2019structured}. Recent studies have explored distilling foundation models into lightweight task-specific networks; however, most assume that teacher and student operate on the same input modality. Cross-modal KD between multiplexed and single-channel biomedical images remains largely unexplored. The proposed framework addresses this gap through cross-modal distillation combined with uncertainty-weighted optimization~\cite{kendall2018}.

\section{Methodology}
The proposed cross-modal knowledge distillation framework
is illustrated in Fig.~\ref{fig:methodology}. The teacher, a frozen foundation model, processes multiplexed images
containing nuclear and membrane channels, generating
pixel-wise soft probability targets. The student, a
lightweight encoder-decoder network, is trained
exclusively on the single nuclear channel, supervised
jointly by the ground truth mask and the teacher's soft
targets. At inference, only the student is deployed with
no dependency on the teacher or multiplexed data.

\subsection{Problem Formulation}

Let $X_m \in \mathbb{R}^{H \times W \times C}$ denote
the multiplexed image with $C$ channels, $X_s \in
\mathbb{R}^{H \times W \times 1}$ the single-channel
counterpart, and $Y \in \{0,1\}^{H \times W}$ the
binary segmentation mask. The objective is to learn a
student model $S(\cdot)$ operating exclusively on $X_s$
that achieves segmentation performance comparable to a
teacher model $T(\cdot)$ with access to the full
multiplexed input $X_m$. This is formulated as
cross-modal KD, where the student learns from both
ground truth masks and teacher outputs to transfer
semantic information from the high-dimensional
multiplexed space into a single-channel representation.

\subsection{Distillation Objective}

The teacher $T(\cdot)$ processes multiplexed input $X_m$
and produces pixel-wise logits $Z_t = T(X_m) \in
\mathbb{R}^{H \times W}$, which are cached to disk prior
to student training. The student $S(\cdot)$ processes
single-channel input $X_s$ and produces logits $Z_s =
S(X_s) \in \mathbb{R}^{H \times W}$. KD is performed by
aligning the pixel-wise sigmoid probabilities of the
teacher and student. Rather than temperature-scaled KL
divergence as in classification KD~\cite{hinton2015distilling},
MSE between sigmoid probabilities is employed, which
is better suited to binary single-logit segmentation
outputs:

\begin{equation}
\mathcal{L}_{distill} = \frac{1}{HW} \sum_{i=1}^{HW}
\left( \sigma(Z_t^i) - \sigma(Z_s^i) \right)^2
\label{eq:distill}
\end{equation}

where $\sigma(\cdot)$ is the sigmoid function. This
formulation directly minimizes pixel-wise probability
discrepancy between teacher and student, providing a
smooth and stable training signal.

\subsection{Segmentation and Boundary Losses}

The supervised segmentation loss combines binary
cross-entropy and Dice loss to jointly optimize
pixel-wise accuracy and overlap under class imbalance:

\begin{equation}
\mathcal{L}_{seg}^{(S)} = \mathcal{L}_{BCE}(S(X_s), Y)
                         + \mathcal{L}_{Dice}(S(X_s), Y)
\label{eq:seg}
\end{equation}

Cell boundary pixels are critical for downstream
morphological analysis and represent the regions
where nuclear-only models most frequently fail. A
boundary-aware loss $\mathcal{L}_{bnd}$ concentrates
gradient signal at these locations. The boundary mask
$M_{bnd}$ is obtained by subtracting the
morphologically eroded ground truth from the original
mask:

\begin{equation}
\mathcal{L}_{bnd} = \mathcal{L}_{BCE}
\left( S(X_s) \odot M_{bnd},\ Y \odot M_{bnd} \right)
\label{eq:bnd}
\end{equation}

where $\odot$ denotes element-wise multiplication.

\subsection{Overall Objective Function}

Balancing segmentation, distillation, and boundary
objectives requires careful loss weighting. Rather
than fixed coefficients, learnable uncertainty
weighting~\cite{kendall2018} is employed to
automatically determine the optimal balance during
training, eliminating manual hyperparameter tuning.
The total loss is defined as:

\begin{equation}
\mathcal{L}_{total} = \frac{1}{2\sigma_1^2}
\mathcal{L}_{seg}^{(S)}
+ \frac{1}{2\sigma_2^2} \mathcal{L}_{distill}
+ \beta\,\mathcal{L}_{bnd}
+ \log(\sigma_1 \cdot \sigma_2)
\label{eq:total}
\end{equation}

where $\sigma_1$ and $\sigma_2$ are learnable scalar
parameters optimized jointly with the student weights.
Scaling each loss term inversely with its associated
uncertainty $\sigma_i^2$ places more weight on
objectives where the model is more confident, while
the logarithmic regularization term prevents the
trivial solution of $\sigma_1, \sigma_2 \to \infty$.
The boundary loss weight $\beta = 0.3$ is fixed, as
boundary pixels constitute a small fraction of the
image and excessive boundary weighting can destabilize
training.

\subsection{Training Procedure}

Training proceeds in two stages. In the first stage,
the teacher processes all training images and caches
pixel-wise soft predictions $\sigma(Z_t)$ to disk.
This decouples teacher inference from student training,
eliminates the memory cost of loading both models
simultaneously, and reduces training time by
approximately $4\times$ compared to live teacher
inference. In the second stage, the student is trained
on single-channel inputs $X_s$ with supervision from
both cached soft targets and ground truth masks $Y$,
according to Eq.~(\ref{eq:total}). The learnable
parameters $\sigma_1$, $\sigma_2$, and all student
weights are optimized jointly. The same two-stage
procedure and distillation objective are applied
identically across all evaluated student architectures;
only the encoder backbone varies between experiments.

\subsection{Cross-Dataset Generalization}

To evaluate robustness under distribution shift,
TissueNet-cached teacher predictions are used to
supervise students trained exclusively on
BBBC038~\cite{caicedo2019nucleus}, without teacher
retraining or target-domain multiplexed data. This
reflects realistic deployment conditions where
multiplexed reference data from the target domain
may be unavailable. The cross-dataset setting is
evaluated across all four student architectures to
assess whether generalization is consistent across
model capacities.
\section{Experimental Setup}

\subsection{Datasets}

\begin{figure}[h!]
    \centering
    \includegraphics[width=\columnwidth]{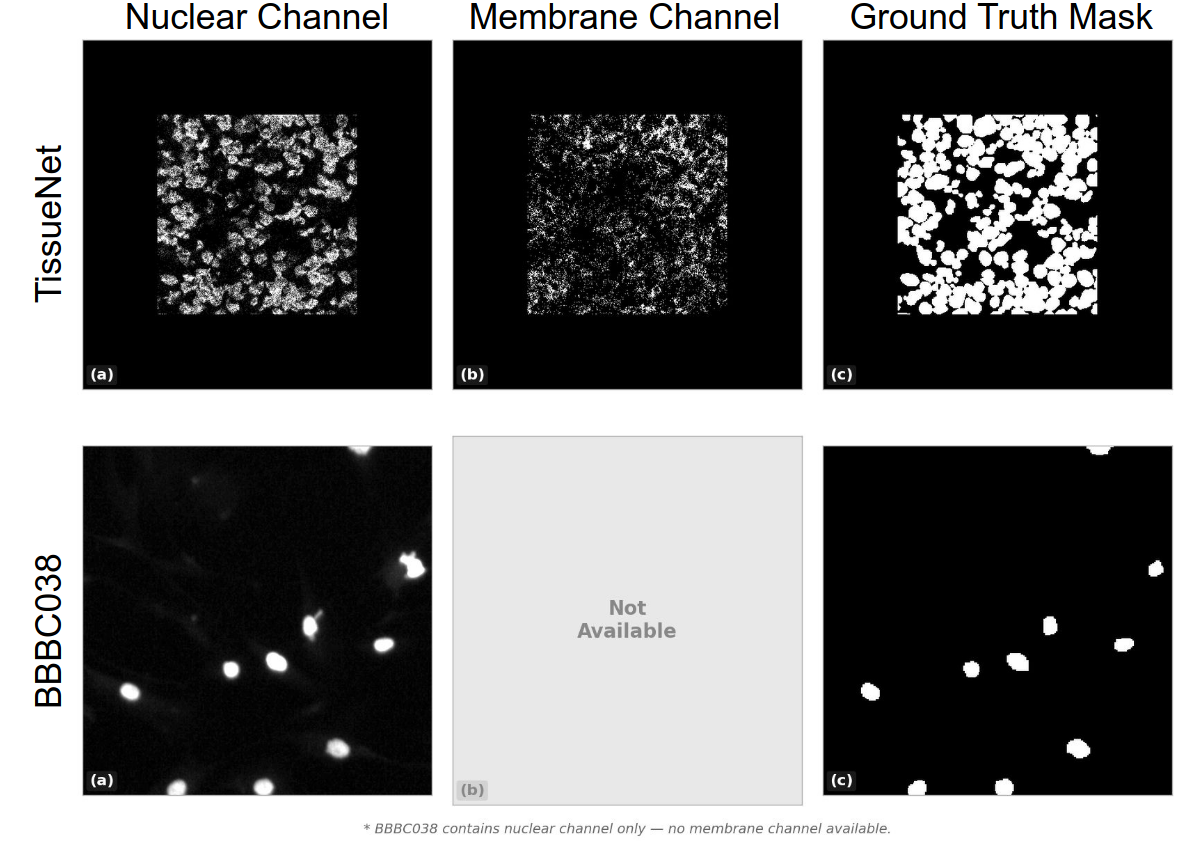}
    \caption{Representative samples from both datasets.
    \textit{Top}: TissueNet provides paired nuclear (a)
    and membrane (b) channels with a binary segmentation
    mask (c). \textit{Bottom}: BBBC038 contains nuclear
    channel only — membrane channel is unavailable (b).
    The teacher receives both TissueNet channels; the
    student receives only the nuclear channel from
    either dataset.}
    \label{fig:datasets}
\end{figure}

Experiments are conducted on two fluorescence microscopy
datasets with different imaging conditions and channel
availability. Representative samples are shown in
Fig.~\ref{fig:datasets}.

\textbf{TissueNet}~\cite{greenwald2022wholecell} is a
large-scale benchmark containing paired nuclear (DAPI)
and membrane channels with corresponding segmentation
masks across diverse tissue types. It serves as the
primary benchmark, where the teacher receives multiplexed
input $X_m$ and the student uses only the nuclear channel
$X_s$ at both training and inference.

\textbf{BBBC038}~\cite{caicedo2019nucleus} (Kaggle 2018
Data Science Bowl) contains 670 single-channel nuclear
fluorescence images from diverse biological contexts.
It is used for cross-dataset evaluation, where
TissueNet-cached teacher predictions supervise students
trained on BBBC038 without teacher retraining, using an
85/15 fixed-seed train-test split.

\subsection{Preprocessing}

All images are center-cropped to $448 \times 448$ pixels.
Each channel is independently normalized by clipping to
the 1st and 99th intensity percentiles and rescaling to
$[0,1]$~\cite{greenwald2022wholecell}. For BBBC038, RGB
images are first converted to grayscale using standard
luminance weights~\cite{poynton2012digital}. Training
images are augmented with random horizontal and vertical
flips and $90^\circ$ rotations. No augmentation is
applied at validation or inference.

\subsection{Models}

The framework is teacher-agnostic and can be paired with any foundation model that processes multiplexed inputs and produces pixel-wise predictions.Two frozen zero-shot teachers are evaluated: SAM ViT-H~\cite{kirillov2023sam} (632M) and CellSAM~\cite{israel2023cellsam} (635M). Students follow a U-Net architecture with
ImageNet-pretrained encoders adapted to
single-channel input. Four backbones are
considered: Swin-Tiny (27M), ResNet18 (11M),
EfficientNet-B0 (5.3M), and MobileNetV3
(1.5M). During inference, only the student
and the nuclear channel are required,
achieving up to $421\times$ parameter reduction. 

\subsection{Implementation Details}

Models are implemented in PyTorch and PyTorch Lightning~\cite{falcon2019pytorch}. Students are trained for 50 epochs using AdamW~\cite{loshchilov2018adamw} with a learning rate of $3\times10^{-5}$, 5\% linear warmup, cosine annealing decay, and gradient clipping of 0.5. Early stopping with a patience of 10 epochs is applied based on validation Dice. A batch size of 4 and mixed-precision (FP16) training are used throughout. All experiments are conducted on a single NVIDIA A100 GPU (80 GB).

\subsection{Evaluation Metrics}

Segmentation performance is measured using Dice
coefficient and Intersection-over-Union (IoU):

\begin{equation}
\text{Dice} = \frac{2|P \cap Y|}{|P| + |Y|}, \quad
\text{IoU}  = \frac{|P \cap Y|}{|P \cup Y|}
\end{equation}

where $P$ is the predicted mask and $Y$ the ground truth.
Results are reported as mean $\pm$ standard deviation
over three independent runs, expressed as percentages.

\subsection{Experimental Protocol}

Experiments evaluate teacher upper bounds, single-channel student baselines, knowledge-distilled students, and cross-dataset generalization. SAM ViT-H and CellSAM are first evaluated on multiplexed TissueNet inputs. Students are then trained on the nuclear channel with and without KD. Distillation is performed using both teachers, yielding eight teacher--student combinations. Cross-dataset generalization is assessed by training students on BBBC038 using TissueNet-cached SAM ViT-H predictions without teacher retraining. Across all experiments, preprocessing and optimization settings remain fixed; only the teacher, student backbone, and use of KD vary.

\section{Results}
\subsection{Quantitative Results}
\begin{table*}[t]
\centering
\caption{Segmentation performance comparison across teacher
models, student architectures, and datasets. Results are
averaged over three independent runs and reported as mean
$\pm$ standard deviation. Best result per dataset in
\textbf{bold}.}
\begin{tabular}{llllllcc}
\hline
\textbf{Dataset} & \textbf{Teacher} & \textbf{Student}
& \textbf{Params} & \textbf{Input} & \textbf{Method}
& \textbf{Dice} & \textbf{IoU} \\
\hline
\multirow{14}{*}{TissueNet}
& SAM ViT-H  & {--}            & 632M  & 2-ch & Teacher Oracle
  & 89.12 $\pm$ 1.21 & 81.21 $\pm$ 1.50 \\
& CellSAM    & {--}            & 635M  & 2-ch & Teacher Oracle
  & 81.33 $\pm$ 1.43 & 73.78 $\pm$ 1.33 \\
\cline{2-8}
& {--}       & Swin-Tiny       & 27M   & 1-ch & Baseline 
  & 65.31 $\pm$ 1.35 & 62.25 $\pm$ 1.47 \\
& CellSAM    & Swin-Tiny       & 27M   & 1-ch & Distilled
  & 73.24 $\pm$ 1.56 & 69.58 $\pm$ 1.71 \\
& SAM ViT-H  & Swin-Tiny       & 27M   & 1-ch & Distilled
  & \textbf{78.36 $\pm$ 1.44} & \textbf{73.29 $\pm$ 1.38} \\
\cline{2-8}
& {--}       & ResNet18        & 11M   & 1-ch & Baseline 
  & 63.84 $\pm$ 1.48 & 60.71 $\pm$ 1.52 \\
& CellSAM    & ResNet18        & 11M   & 1-ch & Distilled
  & 71.12 $\pm$ 1.62 & 67.24 $\pm$ 1.66 \\
& SAM ViT-H  & ResNet18        & 11M   & 1-ch & Distilled
  & 75.81 $\pm$ 1.45 & 71.06 $\pm$ 1.53 \\
\cline{2-8}
& {--}       & EfficientNet-B0 & 5.3M  & 1-ch & Baseline 
  & 62.47 $\pm$ 1.61 & 59.28 $\pm$ 1.66 \\
& CellSAM    & EfficientNet-B0 & 5.3M  & 1-ch & Distilled
  & 69.63 $\pm$ 1.74 & 65.81 $\pm$ 1.79 \\
& SAM ViT-H  & EfficientNet-B0 & 5.3M  & 1-ch & Distilled
  & 74.52 $\pm$ 1.51 & 69.84 $\pm$ 1.63 \\
\cline{2-8}
& {--}       & MobileNetV3     & 1.5M  & 1-ch & Baseline 
  & 58.76 $\pm$ 1.82 & 55.17 $\pm$ 1.91 \\
& CellSAM    & MobileNetV3     & 1.5M  & 1-ch & Distilled
  & 66.18 $\pm$ 1.88 & 61.75 $\pm$ 1.95 \\
& SAM ViT-H  & MobileNetV3     & 1.5M  & 1-ch & Distilled
  & 70.83 $\pm$ 1.79 & 65.94 $\pm$ 1.87 \\
\hline
\multirow{12}{*}{BBBC038}
& {--}       & Swin-Tiny       & 27M   & 1-ch & Baseline 
  & 80.19 $\pm$ 1.43 & 77.42 $\pm$ 1.37 \\
& CellSAM    & Swin-Tiny       & 27M   & 1-ch & Distilled
  & 84.52 $\pm$ 1.26 & 80.76 $\pm$ 1.42 \\
& SAM ViT-H  & Swin-Tiny       & 27M   & 1-ch & Distilled
  & \textbf{87.44 $\pm$ 1.28} & \textbf{83.54 $\pm$ 1.59} \\
\cline{2-8}
& {--}       & ResNet18        & 11M   & 1-ch & Baseline 
  & 78.14 $\pm$ 1.49 & 74.91 $\pm$ 1.45 \\
& CellSAM    & ResNet18        & 11M   & 1-ch & Distilled
  & 82.37 $\pm$ 1.41 & 78.65 $\pm$ 1.52 \\
& SAM ViT-H  & ResNet18        & 11M   & 1-ch & Distilled
  & 85.78 $\pm$ 1.33 & 81.94 $\pm$ 1.47 \\
\cline{2-8}
& {--}       & EfficientNet-B0 & 5.3M  & 1-ch & Baseline 
  & 76.92 $\pm$ 1.58 & 73.36 $\pm$ 1.63 \\
& CellSAM    & EfficientNet-B0 & 5.3M  & 1-ch & Distilled
  & 81.24 $\pm$ 1.47 & 77.28 $\pm$ 1.58 \\
& SAM ViT-H  & EfficientNet-B0 & 5.3M  & 1-ch & Distilled
  & 84.81 $\pm$ 1.36 & 80.84 $\pm$ 1.51 \\
\cline{2-8}
& {--}       & MobileNetV3     & 1.5M  & 1-ch & Baseline
  & 74.28 $\pm$ 1.73 & 70.52 $\pm$ 1.78 \\
& CellSAM    & MobileNetV3     & 1.5M  & 1-ch & Distilled
  & 79.13 $\pm$ 1.66 & 75.21 $\pm$ 1.74 \\
& SAM ViT-H  & MobileNetV3     & 1.5M  & 1-ch & Distilled
  & 82.57 $\pm$ 1.54 & 78.43 $\pm$ 1.67 \\
\hline
\end{tabular}
\label{tab:main}
\end{table*}
Table~\ref{tab:main} presents segmentation performance across all experimental conditions. The student models operate with 1.5M--27M parameters at inference compared to 632M for SAM ViT-H, achieving up to a $421\times$ parameter reduction while requiring only a single input channel.
\begin{figure*}[!htbp]
    \centering
    \includegraphics[width=1\linewidth]{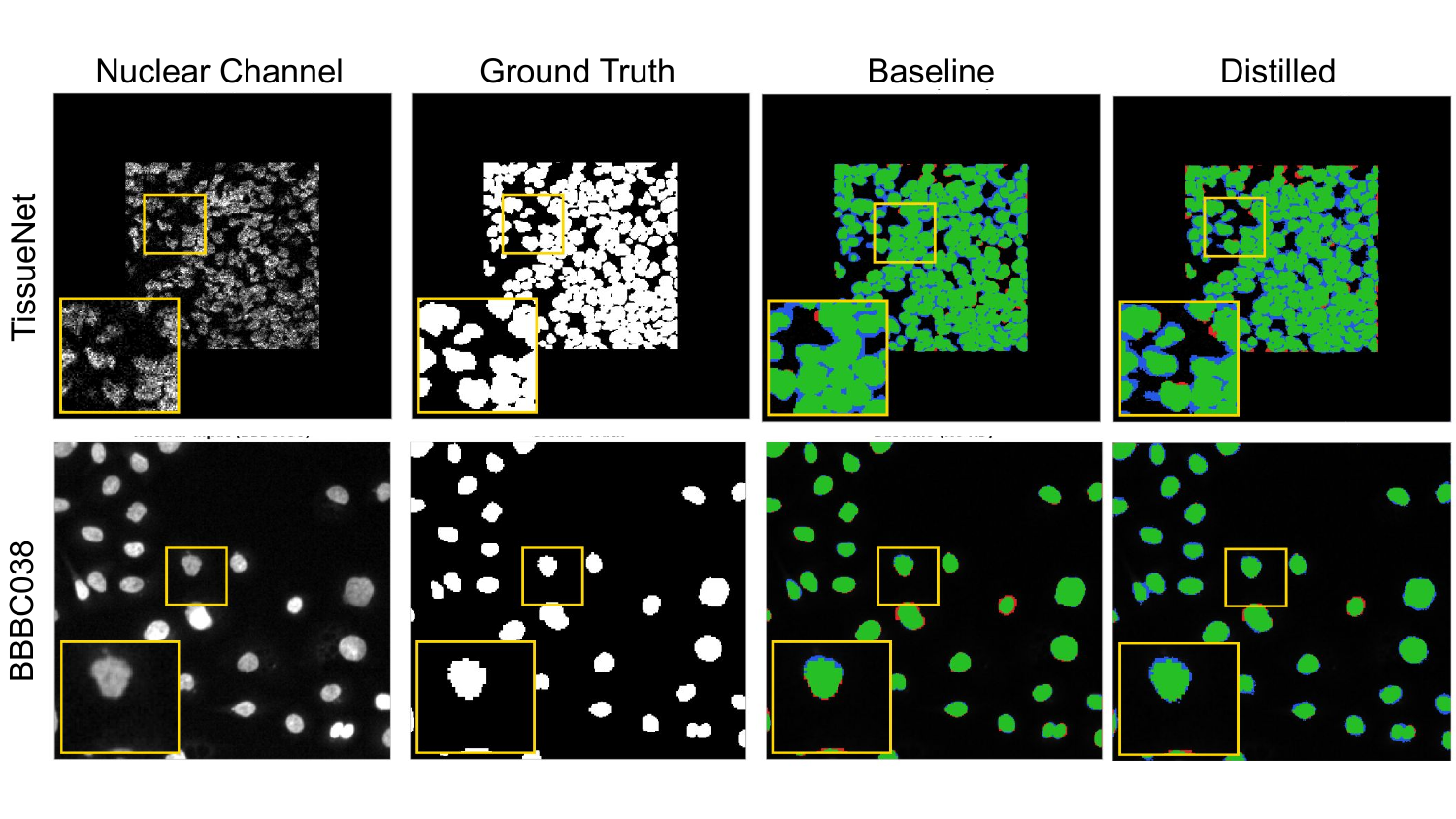}
    \caption{Qualitative segmentation results on TissueNet
    validation set. Each row shows a different sample.
    Columns show nuclear channel input, ground truth mask,
    baseline prediction, and distilled prediction using SAM ViT-H as teacher. Color coding:
    \textcolor[rgb]{0.15,0.75,0.15}{\rule{8pt}{8pt}}~True
    Positive,
    \textcolor[rgb]{0.90,0.15,0.15}{\rule{8pt}{8pt}}~False
    Negative,
    \textcolor[rgb]{0.15,0.35,0.90}{\rule{8pt}{8pt}}~False
    Positive. Yellow boxes indicate magnified inset regions.}
    \label{fig:qual_tissuenet}
\end{figure*}

\subsubsection{Effect of Knowledge Distillation}

KD consistently improves performance over the baseline across all architectures. On TissueNet, SAM ViT-H distillation improves Dice by 13.05, 11.97, 12.05, and 12.07 points for Swin-Tiny, ResNet18, EfficientNet-B0, and MobileNetV3, respectively. The consistency of these gains across architectures spanning an order of magnitude in parameter count confirms that the benefits of distillation are largely independent of model capacity. CellSAM also improves performance over the baseline, although by a consistently smaller margin than SAM ViT-H, indicating that teacher quality directly affects distillation effectiveness.

\subsubsection{Effect of Student Capacity}

Larger student architectures achieve higher absolute performance for both baseline and distilled settings. On TissueNet with SAM ViT-H distillation, Dice scores range from 70.83 for MobileNetV3 (1.5M parameters) to 78.36 for Swin-Tiny (27M parameters), corresponding to a 7.53-point difference across an $18\times$ range in model size. However, the relative gains from KD remain remarkably consistent across architectures (approximately 12 Dice points), suggesting that teacher supervision provides complementary semantic and boundary information regardless of student capacity. Notably, even the smallest student benefits substantially from distillation, supporting deployment in resource-constrained environments.

\subsubsection{Teacher Model Comparison}

SAM ViT-H consistently outperforms CellSAM as a teacher across all students and datasets. On TissueNet, SAM-distilled students exceed CellSAM-distilled students by 4.69--5.12 Dice points, reflecting the 7.79-point difference between the corresponding teacher oracles (89.12 vs. 81.33). This trend suggests that teacher oracle performance on the target domain is a stronger indicator of distillation quality than domain specialization alone.

\subsubsection{Cross-Dataset Results}

On BBBC038, SAM-distilled students improve over baselines by 7.25--8.29 Dice points without teacher retraining, while CellSAM provides smaller but consistent gains of 4.33--4.85 points. These improvements demonstrate that teacher soft-targets capture transferable semantic representations that generalize across imaging conditions and tissue types. The consistent gains across all student architectures indicate that the proposed framework remains effective under cross-dataset distribution shifts.

Overall, the results establish three key findings. First, cross-modal KD consistently closes the modality gap, improving TissueNet performance by approximately 12 Dice points across all student architectures. Second, teacher quality strongly influences distillation effectiveness, with SAM ViT-H consistently producing stronger students than CellSAM. Third, the observed gains on BBBC038 demonstrate that the proposed framework generalizes across datasets without requiring teacher retraining, supporting practical deployment scenarios where multiplexed target-domain data may be unavailable.

\subsection{Qualitative Results}

Qualitative results for the best-performing configuration (SAM ViT-H teacher and Swin-Tiny student) are shown in Fig.~\ref{fig:qual_tissuenet}. The top row presents TissueNet examples, while the bottom row shows cross-dataset results on BBBC038. The color-coded overlays indicate true positives (green), false negatives (red), and false positives (blue).

On TissueNet, the baseline model produces fragmented predictions with noticeable boundary errors, particularly in densely packed regions where nuclear signals alone are insufficient to delineate neighboring cells. In contrast, the distilled model recovers more coherent cell boundaries and reduces both false positives and false negatives. The zoomed regions further demonstrate that distillation is especially effective in crowded cellular areas, where membrane information available to the teacher provides additional structural guidance.

On BBBC038, segmentation is generally easier because nuclei are larger and more isolated. Nevertheless, the distilled model still produces cleaner boundaries and fewer segmentation errors than the baseline model. The consistent improvements observed across both datasets visually support the quantitative gains reported in Table~\ref{tab:main} and indicate that the transferred knowledge remains effective despite the distribution shift between TissueNet and BBBC038.
\section{Conclusion}

This work presented a cross-modal knowledge distillation framework for tissue segmentation under missing-channel constraints. A frozen multiplexed teacher transfers information from nucleus and membrane channels to lightweight single-channel students. This enables accurate segmentation using only the nucleus channel during inference. Across four student architectures and two datasets, knowledge distillation consistently improved segmentation performance while reducing computational complexity. SAM ViT-H also produced stronger student performance than CellSAM. It suggests that teacher quality plays an important role in effective knowledge transfer. Cross-dataset results on BBBC038 further showed that the framework generalized well without requiring teacher retraining. This highlights its potential for practical biomedical imaging applications where multiplexed data may not always be available. Future work will extend this framework to instance and panoptic segmentation, multi-teacher distillation, and additional modalities such as H\&E histology and phase-contrast microscopy.
\balance
\bibliographystyle{IEEEtran}
\bibliography{references}

\end{document}